\documentclass[10pt,twocolumn,letterpaper]{article}

\usepackage{ijcb}
\usepackage{times}
\usepackage{epsfig}
\usepackage{graphicx}
\usepackage{amsmath}
\usepackage{amssymb}
\usepackage{stfloats}

\ijcbfinalcopy 

\ifijcbfinal\fi
\begin{document}

\title{UMSNet: An Universal Multi-sensor Network for Human Activity Recognition}

\author{Jialiang Wang\\
\and
Haotian Wei\\
\and
Yi Wang\\
\and
Shu Yang\\
\and
Chi Lin\\
}

\maketitle
\thispagestyle{empty}

\begin{abstract}
   Human activity recognition (HAR) based on multimodal sensors has become a rapidly growing branch of biometric recognition and artificial intelligence. However, how to fully mine  multimodal time series data and effectively learn accurate behavioral features has always been a hot topic in this field. Practical applications also require a well-generalized framework that can quickly process a variety of raw sensor data and learn better feature representations. This paper proposes a universal multi-sensor network (UMSNet) for human activity recognition. In particular, we propose a new lightweight sensor residual block (called LSR block), which improves the performance by reducing the number of activation function and normalization layers, and adding inverted bottleneck structure and grouping convolution. Then, the Transformer is used to extract the relationship of series features to realize the classification and recognition of human activities. Our framework has a clear structure and can be directly applied to various types of multi-modal Time Series Classification (TSC) tasks after simple specialization. Extensive experiments show that the proposed UMSNet outperforms other state-of-the-art methods on two popular multi-sensor human activity recognition datasets (i.e. HHAR dataset and MHEALTH dataset).
\end{abstract}

\section{Introduction}

Nowadays, people tend to utilize artificial intelligence (AI) to make their lives safe, health and smart. Human activity recognition (HAR) is an important basis for many practical applications, such as medical care \cite{medical}, emotional response based on heart rate \cite{emotional-a,emotional-b}, human activity monitoring \cite{Human}, and health tracking \cite{tail}. HAR based on image data has been widely studied, but it has inevitable disadvantages: the limitation of usage scenarios, the issue of infringing personal privacy, and high requirements on the performance and storage of the image capture equipments. HAR based on time series data from wearable sensors can overcome the above problems. Many sensors have been integrated into mobile devices, such as mobile phones and watches, which can collect user data anytime and anywhere. In addition, most of the data collected by sensors are physical quantities such as acceleration and angular velocity, which have little infringement on personal privacy. 

Human activity recognition (HAR) based on multi-modal wearable sensors is a challenging multivariate time series classification problem (TSC)\cite{wang2019deep}. Although many deep-learning based works have been done to solve multi-modal feature extraction and fusion, spatial-temporal feature extraction, and classification problems involved in HAR, they still have limitation in real-world scenarios \cite{wang2019deep,ebrahim2020quantitative,fawaz2020deep}. What is needed for practical applications is to build a framework with good generalizability that can quickly process raw sensor data and learn better feature representations.

To address the above issues, we propose a universal deep learning framework (UMSNet) for multi-modal time series classification. The core of the UMSNet is the integration of the residual network\cite{FCNN} and the Transformer\cite{transformer}. The residual network is used to encode local and global multi-modal features for sensors. The Transformer is used to learn spatial-temporal features. Particularly, to improve the performance of the residual network, we design a new and efficient lightweight sensor residual (LSR) block. LSR block reduces the number of activation function and normalization layers and uses an inverted bottleneck structure and grouping convolution. The entire framework can be easily customized with the Residual Net part and the Transformer part for a variety of identification tasks. 

Specifically, the main contributions of this paper are as follows:
\begin{itemize}
\item Framework contribution: A universal multi-sensor learning (UMSNet) framework is proposed. The UMSNet is composed of the residual network and the Transformer, which can be applied to various multi-modal time series classification (TSC) tasks. 

\item Network contribution: An efficient and lightweight sensor residual (LSR) block is proposed to fuse multi-modal features. LSR block improves the performance by reducing the number of activation layers and normalization layers, and adding inverted bottleneck structure and grouping convolution.Use layerScale  and stochastic depth as training strategies.

\item Empirical contribution: The UMSNet achieves new state-of-the-art performance on two benchmark datasets(i.e., HHAR dataset and MHEALTH dataset).
\end{itemize}

\section{Related Works}
In this section, we briefly review the residual network, Transformer-based network, multi-modal network, human activity recognition, and related literatures. For more details, please refer to surveys\cite{wang2019deep,fawaz2020deep,ebrahim2020quantitative}.

\subsection{Residual Network}

With the raising and wide applications of deep-learning technology, classification models based on convolutional neural networks (CNNs) and their variants (e.g., fully convolutional neural networks (FCNs)\cite{FCNN}) have become the mainstream. VGGNet \cite{vgg} improves the performance by increasing the number and depth of network layers on top of previous convolutional neural network (CNN) architectures. ResNet \cite{resnet} builds on VGGNet by explicitly fitting a residual map implemented by a feed-forward neural network combined with shortcut connections. Then increase the network depth to obtain improved results. A remarkable aspect of another well-known network, GoogleLeNet \cite{googlelenet}, is that its architecture greatly improves the utilization of computing resources. In a well-designed network, the computational overhead remains the same as the model depth and width increase. SENet \cite{senet}, the winner of the ILSVRC 2017 classification category, recalibrates the channel-wise feature responses without relying on new spatial dimensions. EfficientNet \cite{efficientnet}, a multidimensional hybrid model scaling method, greatly improves the training speed of the network. Regnet \cite{regnet}, like hand-designed networks, aims at interpretability, which can describe some general design principles for simple networks and generalize them in various settings. ConvNeXt \cite{convnext} studies the architectural differences between ConvNets and Transformers, and identify confounding variables when comparing network performance. However, the design of residual network is not refined enough for the fusion of multimode data, and there are many parts that can be optimized.

\subsection{Transformer-based Networks}

The Transformer \cite{transformer} has achieved great success in the field of artificial intelligence. Improvements to Transformer are also emerging. Bert \cite{bert} emphasizes that the traditional one-way language model or the method of shallowly splicing two one-way language models for pre-training is no longer used as before, but the new masked language model (MLM) is used for pre-training, so that deep bidirectional linguistic representations can be generated. Iz Beltagy et al. \cite{longformer} propose a sparse attention mechanism that uses a combination of local self-attention and global self-attention (or sparse attention for short), and optimize it with CUDA to maximize the model Capable of accommodating tens of thousands of texts in length while still achieving better results. Google proposed the Vision Transformer (ViT) \cite{vit}, which can directly use Transformers to classify images without the need for convolutional networks. The ViT model achieves results comparable to current state-of-the-art convolutional networks, but requires significantly less computational resources for its training. MSRA propose a new visual Transformer called Swin Transformer \cite{swin}, which can be used as a general backbone for computer vision, bringing the challenge of adapting Transformer from language to vision. However, using the Transformer alone, it is difficult to adapt to sensor data in a variety of data formats and requires a very large training set to achieve good results\cite{han2021transformer}. 

\subsection{Multi-modal Network}
QA Zhen et al. \cite{imaging} propose to encode the time series of sensor data into images (i.e., encode a time series into a two-channel image) and use these transformed images to preserve the features required for human activity recognition. The Temporal Fusion Transformer (TFT) \cite{tft} proposes interpretable deep learning for time series forecasting. A novel attention-based architecture that combines high-performance multi-level forecasting with an interpretable understanding of temporal dynamics. Google proposed XMC-GAN \cite{xmc-gan}, a cross-modal contrastive learning framework to train GAN models for text-to-image synthesis, for research addressing the problem of generated cross-modal contrastive loss. ALIGN \cite{align} (A Large-scale Image and Noisy-Text Embedding) model, designed to address the problem that current vision and visual language SOTA models rely heavily on specific training datasets that require expert knowledge and extensive labels. 

\subsection{Human Activity Recognition}
Deepsense \cite{deepsense} proposes a deep learning framework that runs on end devices, which can locally acquire sensor data that needs to be processed and apply deep learning models, such as convolutional neural networks, to these data without uploading to the cloud or gated recurrent neural networks. Bianchi V et al. \cite{bianchi2019iot} combine wearable devices with deep learning technology to propose a new human motion recognition system, the system uses convolutional neural networks to classify data on 9 different behaviors of users in a home environment and achieve a good classification effect. To characterize the relevance of the data in space, T-2DCNN and M-2DCNN convolutional network model \cite{deng2019convolutional} encoded the three-axis data of the inertial sensor into a data picture format, and then extracted the spatial correlation between the sensors and the temporal synchronization correlation characteristics of the three-axis data through convolution operation. Tufek N et al. \cite{tufek2019human}  developed a human motion recognition system with using accelerometers and gyroscopes. The system utilizes a modified LSTM network to develop a behavioral classification system based on multi-type sensor data.

\section{Methodology}

\begin{figure*}[!htb]
\begin{center}

\includegraphics[height=12.5cm]{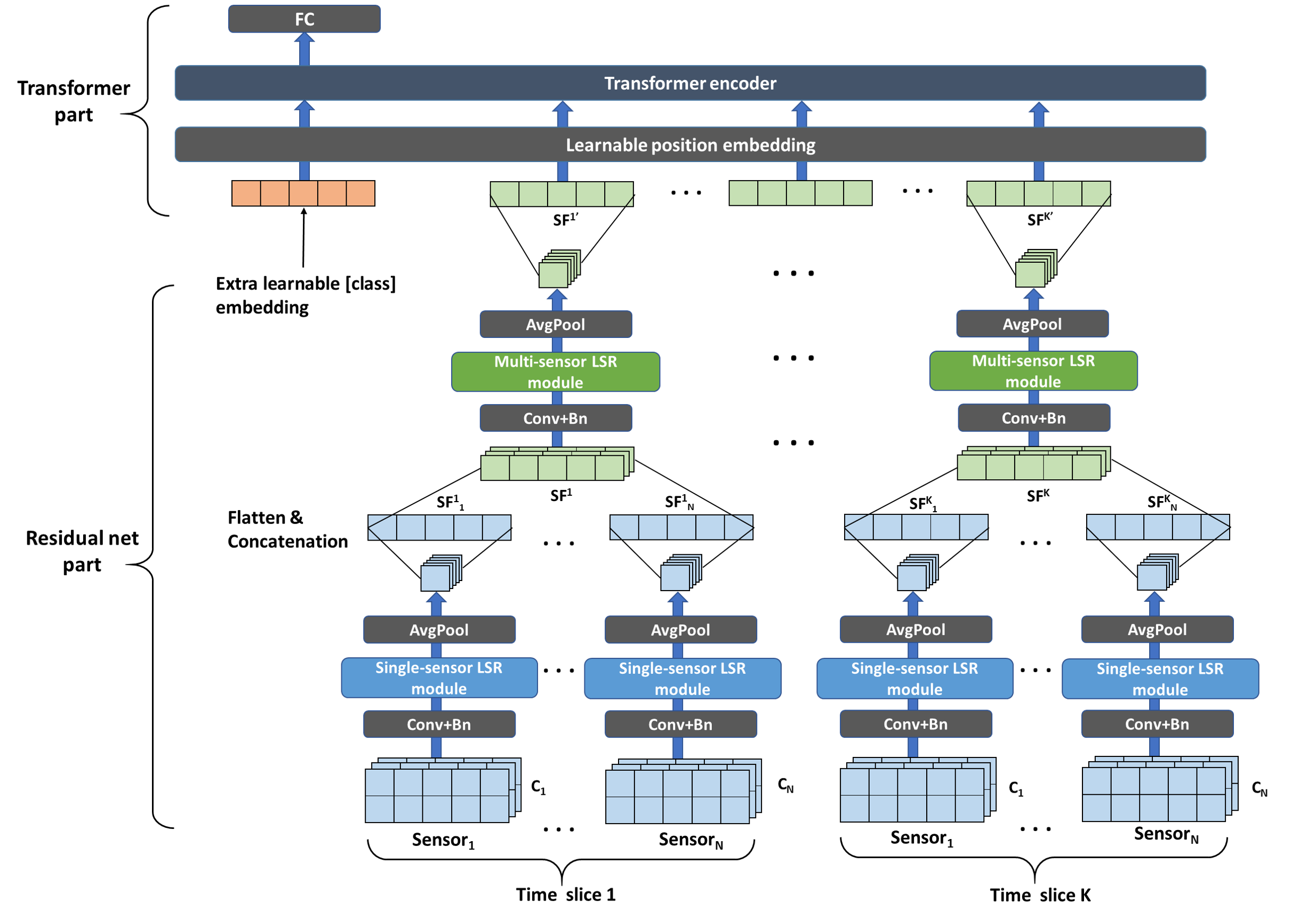}

\end{center}
  \caption{The structure of UMSNet. It is mainly composed of a Residual net part and a Transformer part. Lightweight sensor residual (LSR) module is the
  core of the Residual net part. The LSR Module consists of LSR blocks.
  Given a time series, the UMSNet processes it from bottom to top in three stages: single-sensor  feature  extraction, multi-sensor feature fusion and extraction, and multi-sensor sequence feature extraction and classification. }
\label{fig:short}
\end{figure*}

\subsection{Overview}
In this section, we present the details of the proposed UMSNet. The network structure of UMSNet is shown in Figure 1. Given a time series $X$, the UMSNet processes it from bottom to top in three stages. 

Stage 1: Single-sensor feature extraction. The proposed Lightweight sensor residual (LSR) module is used to carry out convolution operation to extract features, forming a unified sensor feature ($SF_i$) for the sensor $i$.  

Stage 2: Multi-sensor feature fusion and feature extraction; In this stage, the features from $N$ sensors are flattened and concatenated, and be taken as channels of a feature map of all sensors, named $SF^K$ in each time slice $K$. 

Stage 3: Multi-sensor sequence feature classification. The global information of $N$ channels is merged into the $SF^{K'}$ and be input to the Transformer to get the classification results.  




\subsection{Lightweight Sensor Residual (LSR) Block}

\begin{figure}[h]
\begin{center}
\includegraphics[height=6cm]{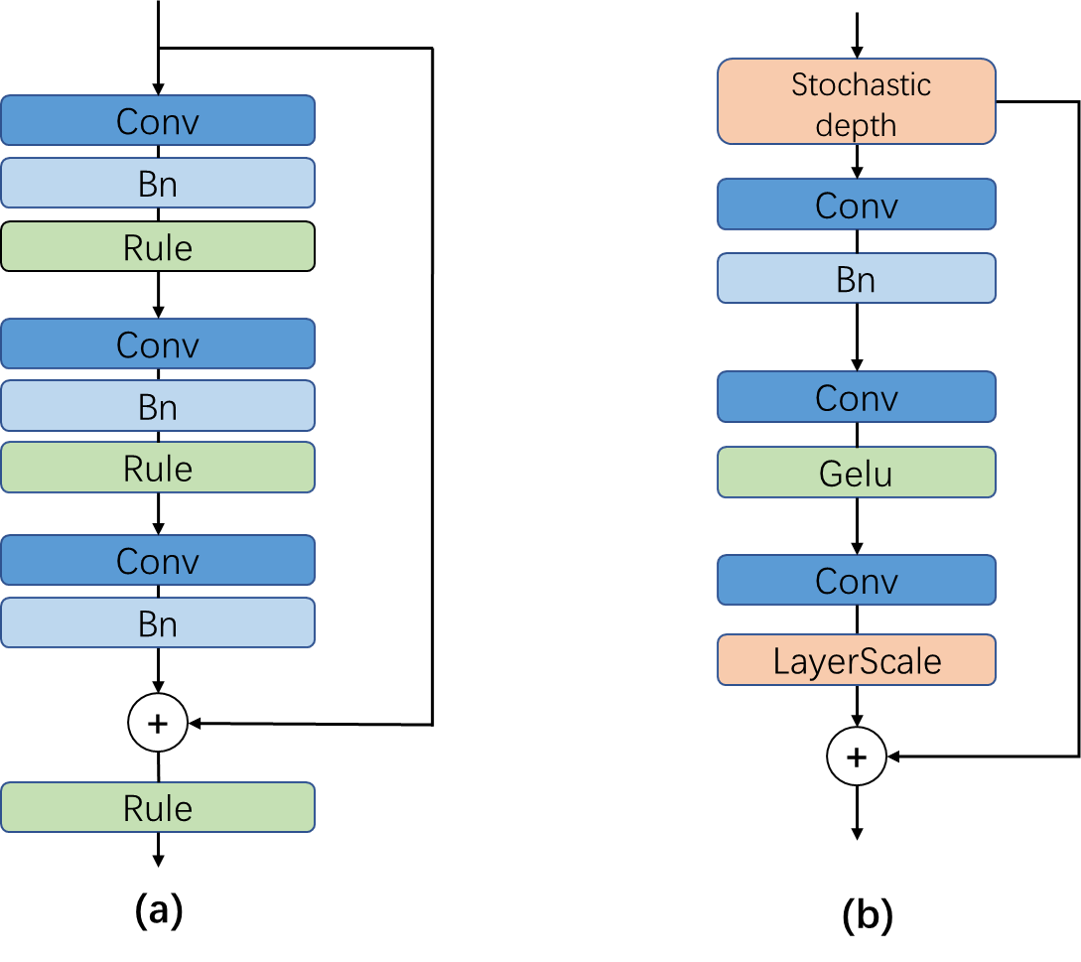}
\end{center}
   \caption{Illustration of the proposed Lightweight Sensor Residual (LSR) Block and the Resnet residual block\cite{resnet}. (a): A Resnet residual block structure.
            (b): A LSR block structure.}
\label{fig:short}
\end{figure}

\begin{figure}[h]
\begin{center}
\includegraphics[height=6cm]{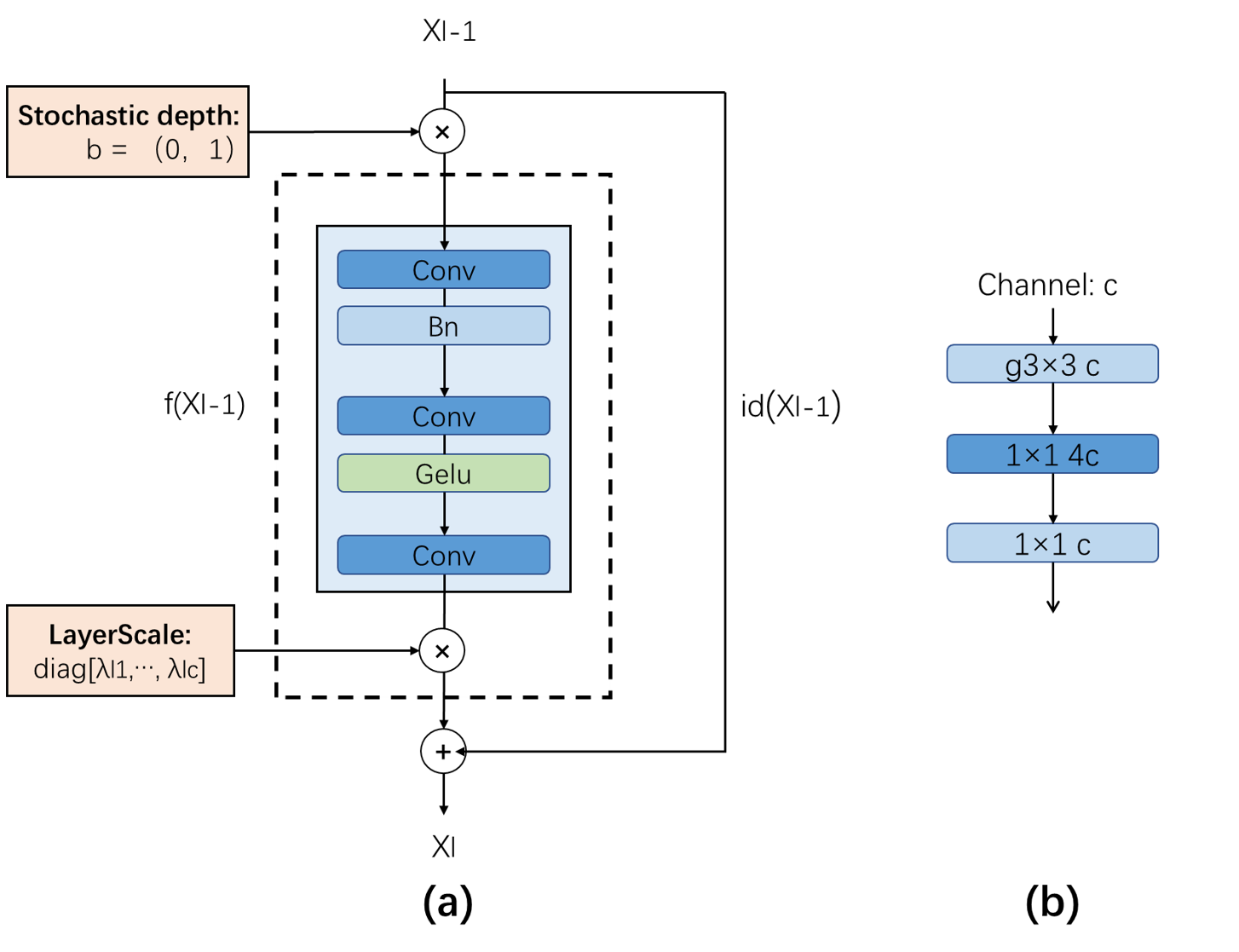}
\end{center}
   \caption{ Details and an example of the LSR block.
   (a): Details of LayerScale and Stochastic depth in the LSR block.
   (b): An example of a channel change after convolution, where $c$ is channel number, [g$3\times3$] represents the grouping convolution layer with a [$3\times3$] convolution kernel, and [$1\times1$] represents the convolution layer with a [$1\times1$] convolution kernel. 
 }
\label{fig:short}
\end{figure}



In order to improve the efficiency of multi-modal feature extraction of time series, we design a lightweight sensor residual block, called LSR block, by modifying the original residual network block structure. The LSR block is the core of the Single-sensor LSR module and the Multi-sensor LSR module in the UMSNet. The specific improvements are as follows:
\begin{itemize}
\item We first reduce the number of activation functions and the number of normalization layers. As the ConvNeXt \cite{convnext} network proves that too frequent nonlinear projections are actually detrimental to the information transfer of network features. So it is wise to use fewer activation functions and normalization in a residual block. The regular Resnet block uses three normalization layers and three activation functions, as shown in Figure 2(a). Here, in our new residual block, we only use one normalization layer and one activation function, as shown in Figure 2(b).
\item Inspired by MobileNetV2 \cite{mobilenet}, we adopt the inverted bottleneck structure in LSR block. Our LSR block in the form of small-large-small structure, as shown in Figure 3(b). In this way, information can be converted between feature spaces of different dimensions, avoiding information loss caused by dimension compression. Specifically, we expand the dimension of the input data to four times the original size in the second layer of convolution in the residual block. In the third layer of convolution, the dimensions are restored to the original size.
\item We also replace the ordinary $[3\times3]$ convolutional layer with grouped convolution to make the network more lightweight. We use the maximum number of groups (the number of groups and the number of channels are the same), so that the number of parameters of the grouped convolutional layer is $\frac{1}{c} $ the public convolutional layer ($c$ is the number of channels). This form of grouped convolution is similar to the self-attention mechanism, in which spatial information is mixed in a single channel because each convolution kernel processes a channel separately.  
\end{itemize}
 
In addition, We use LayerScale \cite{layerscale} and Stochastic depth \cite{stochastic} as LSR block training strategies, as shown in Figure 3(a).  

\textbf{LayerScale:} In the LSR block, different channels are multiplied by a parameter that can be learned to make features more refined and accurate. This process can be represented by the following equation:  
\begin{equation}
x_{residual} = f(x_{l-1})\times diag(\lambda _{l1},...,\lambda_{lc})
\label{3}
\end{equation}
where $f$ represents the residual part, $X_{residual}$ is the output of the residual link for this block, $x_{l-1}$ is the output of the last LSR block, and $\lambda _{l1},...,\lambda_{lc}$ are learnable parameters.  

\textbf{Stochastic depth}: We use the stochastic depth to improve the generalization capability of the model. During training, a random variable $b$ is added, satisfying the ($0-1$) distribution, $x_{residual}$ is multiplied by $b$, and the residual part was randomly discarded. This process can be expressed by the following equation: 
\begin{equation}
x_{l} = id(x_{l-1}) + bx_{residual}
\label{3}
\end{equation}
where $id$ represents the identity mapping, $X_{residual}$ is the output of the residual link for this LSR block, $x_{l}$ is the output through this LSR block,$x_{l-1}$ is the output of the last LSR block.

\subsection{Transformer part}

The Transformer part of UMSNet consists of Learnable Position embedding\cite{vit} and Transformer Encoder\cite{transformer}. The Transformer part aims to extracting the sequence feature of sequence data.  

The Transformer \cite{transformer} is a model based on self-attention mechanism, which achieves higher accuracy and performance than conventional RNN. Here, we only use The Transformer Encoder to learn time series features. Before the Transformer Encoder, we add a learnable position embedding to each slice feature ($SF'$) as position information. Meanwhile, inspired by Transformer-based Bert model \cite{bert}, we insert a specific classification token ([class]) at the beginning of the time series. Both [class] and Position embedding are parameters of the network and can be learned and optimized automatically. The [class] token from the last Transformer layer is used to aggregate the entire series representation information. Finally,  ${Fc}$ (full connection layer) is used for classification.  

Also, like Residual net part, you can customize the Transformer part with different parameters and depths depending on the task needs.  

\section{Experiments}
\subsection{Datasets}
The followingdata sets are used for evaluating the performance of the proposed method.

\textbf{HHAR}\cite{hhar} is a data set designed to serve as a benchmark for human activity recognition algorithms (classification, automatic data segmentation, sensor fusion, feature extraction, etc.) in real-world environments. Specifically, HHAR contains the readings from two motion sensors (i.e., the accelerometer and the gyroscope) in the smartphones and the smartwatches. This dataset contains six activities (i.e. cycling, sitting, walking, standing, walking upstairs, and walking downstairs) from nine users and six mobile devices (four smartphones and two smartwatches).  

\textbf{MHEALTH}\cite{mhealth} is a mobile health data set based on multimodal wearable sensor data for the human activity recognition. Sensors are placed on the subjects' chest, right wrist and left ankle to measure movement of the body, using acceleration, angular velocity and magnetic field direction signals. In addition, the sensor placed on the chest provides two sets of lead to provide 2-lead Electrocardiograph measurements. Human activity in MHEALTH can be divided into seven categories, namely: standing, sitting and relaxing, lying down, walking, climbing stairs, cycling, and jogging.

\subsection{Metrices}

Since human activity recognition is a classification problem, each class can be regarded as an independent set of samples for the positive class, and the other classes for the negative class. Therefore, the classification performance is generally assessed by the Accuracy and Macro-F1 scores. The computational complexity of the network is evaluated by the Params, Mult-Adds and Time.  

\textbf{Accuracy} \cite{imaging} is the proportion of correctly identified examples in all samples and can be defined by
\begin{equation}
Accuracy = \frac{1}{n_{samples}} \sum_{i=1}^{n_{samples}} \iota (\hat y_{i} = y_{i}),  \label{1}
\end{equation}
where $\hat y_{i}$ is the predicted value of the $i^{th}$ sample, $y_{i}$ is the corresponding true value, and $\iota$ is the indicator function.

F1-score is an index used to measure the accuracy of multi-classification models in statistics, which is the harmonic mean of the Precision and the Recall. F1-score is defined by
\begin{equation}
F1_i=2\frac{Recall_{i}\times Precision_{i}}{Recall_{i}+Precision_{i}}
\label{2}
\end{equation}

\textbf{Macro-F1} \cite{imaging} is suitable for multi-classification problems and is not affected by data imbalance. Given $n$ class, $i\in (1,...,n)$, Macro-F1 is defined by
\begin{equation}
MacroF1=\frac{\sum_{i=1}^{n} F1_i}{n} 
\label{3}
\end{equation}

\textbf{Params} refers to the number of parameters that can be learned and optimized in the network.

\textbf{Multi-Adds} indicates the number of arithmetic operations such as multiplication and addition on the network.

\textbf{Time} is needed to identify a sample of human activity(base on RTX3080).

\subsection{Baseline Methods}

We extensively evaluate the performance of the proposed UMSNet with several widely used architectures in HAR:  

LSTM \cite{lstm}: LSTM is a special RNN, which is mainly used to solve the problems of gradient disappearance and gradient explosion in the process of Long series training. 
In the experiments, we use the three-layer LSTM. 

GRU \cite{gru} : GRU is an effective variant of the LSTM network, and its structure is simpler than that of LSTM network. In the experiments, we use the three-layer GRU.  

Resnet \cite{resnet} : Resnet use residual structure to solve the degradation problem, allows the network to deepen as much as possible.  In the experiments, we use the Resnet50.

Efficientnet \cite{efficientnet} : Efficientnet can improve the index and reduce the number and calculation of model parameters, by comprehensively optimizing network width, network depth and resolution. In the experiments, we use the Efficientnet-B1.  

Regnet \cite{regnet} : Regnet parameterizes the group of the network (e.g., Convolutional LSTMs or Convolutional GRUs), which are shown to be good at extracting Spatio-temporal information. In the experiments, we use the Regnet50.


Since the parameters of the LSR can be adjusted according to the needs of tasks and different sensor data formats and residual networks of different depths can be built, we design three kinds of the UMSNet with different structures in terms of the depths of the network for experiments as shown in Table 1, to demonstrate the flexibility of our network. 

\begin{table*}[!ht]
    \begin{center}
    
    \begin{tabular}{|c|c|c|c|}
    \hline
        \textbf{Depth} & \textbf{Single sensor LSR module} & \textbf{Multi-sensor LSR module} & \textbf{Transformer encoder}  \\ \hline\hline
        UMSNet-A & [2, 2, 2, 2] & [2, 2, 2, 2] & 3  \\ \hline
        UMSNet-B & [2, 2, 6, 2] & [2, 2, 6, 2] & 6  \\ \hline
        UMSNet-C & [2, 2, 18, 2] & [2, 2, 18, 2] & 6  \\ \hline
    \end{tabular}
    \end{center}
    \caption{The numbers in the table represent the quantity of the LSR blocks. For example, [2,2,2,2] represents a total of 8 LSR blocks, with a down sampled layer inserted between each two Residual blocks. }
\end{table*}

\begin{table*}[!htb]
    \begin{center}
        
    \begin{tabular}{|c|c|c|c|c|c|c|}
    \hline
        \textbf{Method} &\multicolumn{2}{|c|}{$K=6$} & \multicolumn{2}{|c|}{$K=12$}  &  \multicolumn{2}{|c|}{$K=24$}    \\ \cline{2-7}
       \textbf{ Comparison} & \textbf{Accuracy} & \textbf{Macro-F1} & \textbf{Accuracy} & \textbf{Macro-F1} & \textbf{Accuracy} & \textbf{Macro-F1}  \\ \hline\hline
        GRU\cite{gru} & 0.7858 & 0.7932 & 0.8378 & 0.8303 & 0.8474 & 0.8287  \\ \hline
        LSTM \cite{lstm}& 0.8093 & 0.8008 & 0.8398 & 0.8299 & 0.8564 & 0.8542  \\ \hline
        ResNet\cite{regnet} & 0.8815 & 0.8701 & 0.8936 & 0.8798 & 0.8808 & 0.8559  \\ \hline
        EfficientNet\cite{efficientnet} & 0.8991 & 0.9039 & 0.8964 & 0.8778 & 0.9052 & 0.8939  \\ \hline
        Regnet\cite{regnet} & 0.9079 & 0.9197 & 0.9079 & 0.9085 & 0.9166 & 0.9132  \\ \hline
        UMSNet-A(Ours) & 0.9155 & 0.9093 & 0.9251 & 0.9312 & 0.9329 & 0.9342  \\ \hline
        UMSNet-B(Ours) & 0.9171 & 0.9181 & \textbf{0.9304} & \textbf{0.9327} & 0.9358 & 0.9382  \\ \hline
        UMSNet-C(Ours) & \textbf{0.9187} & \textbf{0.9211} & 0.9289 & 0.9304 & \textbf{0.9397} & \textbf{0.9402}  \\ \hline
    \end{tabular}
    \end{center}
    
    \caption{Performance of different methods on HHAR dataset. $K$ is the slice number of a time series data.}
\end{table*}

\begin{table}[!h]
\begin{center}
    \begin{tabular}{|c|c|c|c|}
    \hline
        \textbf{Method} & \textbf{Params} & \textbf{Multi-Adds} & \textbf{Time(ms)} \\ \hline\hline
        GRU\cite{gru} & 17712 & 17280 & 7.122 \\ \hline
        LSTM \cite{lstm}& 23616 & 23040 & 10.023  \\ \hline
        ResNet \cite{regnet} & 25.557M & 4.089G  & 93.053\\ \hline
        Efficientnet\cite{efficientnet} & 7.794M & 569.682M & 87.959 \\ \hline
        Regnet\cite{regnet} & 54.278M & 15.940G & 172.442 \\ \hline
        UMSNet-A & 2.905M & 72.863M & 33.864 \\ \hline
        UMSNet-B & 4.887M & 76.564M & 58.182
 \\ \hline
        UMSNet-C & 6.092M & 82.944M & 79.015
 \\ \hline
    \end{tabular}
    \end{center}
    \caption{Comparison of network parameter, computational cost, and the speed of processing a time series of 6 seconds on HHAR dataset.}
\end{table}

\subsection{Implementation details}

Theoretically, the time length of each slice can be unequal in our UMSNet, which can be tuned according to the practical requirements. For a fair comparison, the time length of each slice is set to be equal for all methods. We divide the sequence data of two datasets into three time lengths of 1.5 seconds, 3 seconds, and 6 seconds, and we take 0.25s as a slice time length, so a time series $X$ has three different sample numbers $K=6,12,24$. 

We used Gelu as the activation function to be consistent with Transformer.  The down sampled layer between different residual blocks is composed of $Bn$ plus a [$2\times2$] convolution with $stride=2$.  

\subsection{Experimental results on HHAR dataset}

The HHAR data set has two types of sensors, an acceleration sensor and a gyroscope(angular velocity sensor). The $x$, $y$, and $z$ axes of each sensor are taken as the three channels of input data.  

For all methods, the leave-one-user-out test method is used, that is, for the 9 users in the HHAR data set, 8 users are taken as the training set and another user is taken as the test set to test each user. Experimental results are presented in Table 2, Figure 4 and Table 3.

\begin{figure}[!htb]

\begin{center}

\includegraphics[height=8cm]{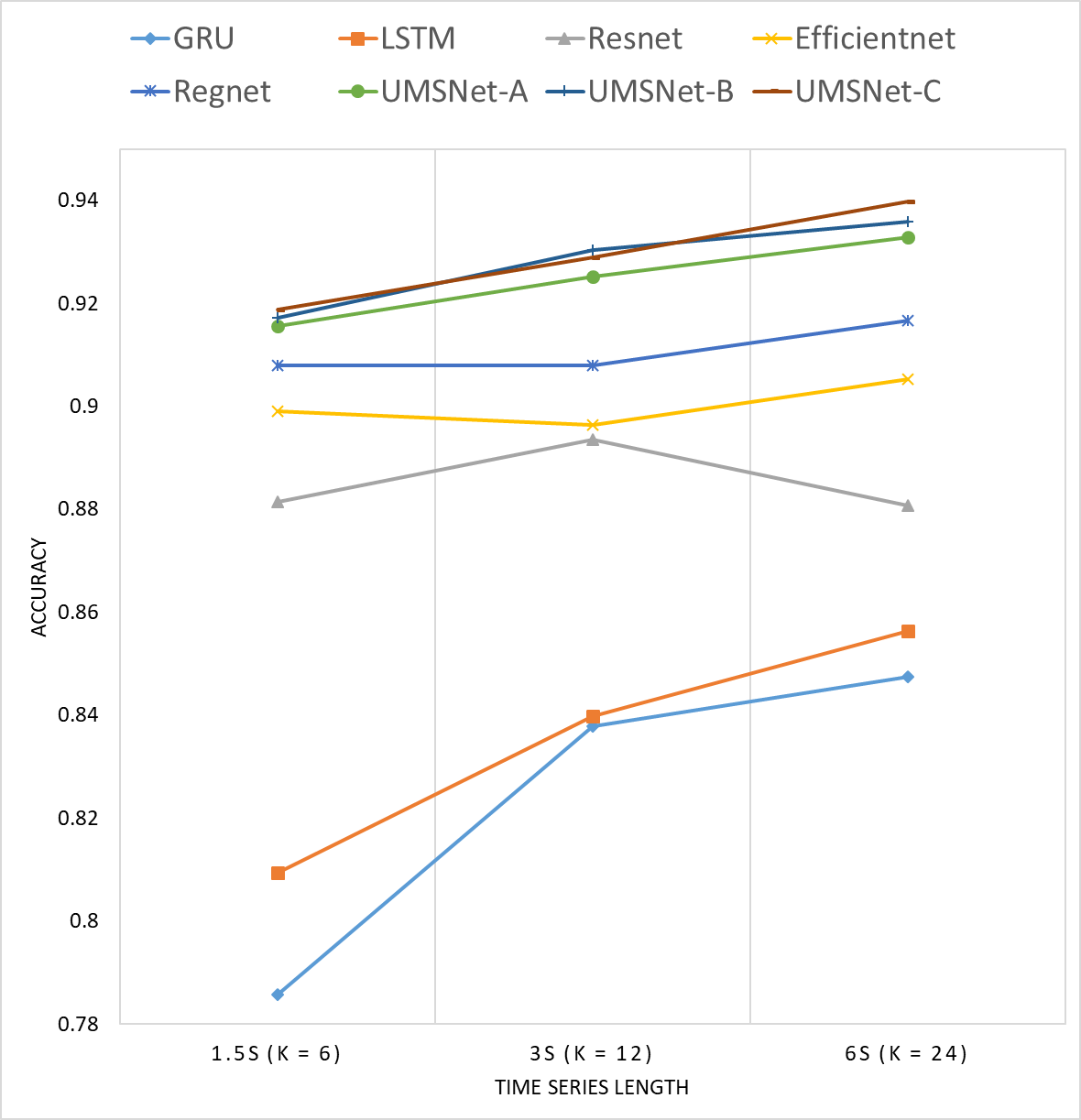}

\end{center}
    \caption{Accuracy of different methods of three time series length setting on HHAR dataset.}
\label{fig:short}
\end{figure}

Experimental results show that our methods are superior to other SOTA methods in terms of accuracy and macro-F1 over three-length settings of times series. It can be seen from the accuracy curves (in Figure 4) of the three time slice lengths that the recognition rate of all algorithms increases with the increase of the sample number of a time series. In particular, recognition rate of the three structures of our UMSNet increases stably, indicating the stability of our network designs. Other methods, such as Resnet and Efficientnet, are all irregular and unstable.

In addition, the network computation cost of UMSNet is far less than the conventional residual network based classification network, i.e., ResNet\cite{resnet} and Regnet\cite{regnet}, in Table 3. Although the network parameters and computation cost of the UMSNet are larger than that of the simple RNN networks, i.e., GRU\cite{gru} and LSTM\cite{lstm}, the accuracy of the UMSNet is far better than that of them. The UMSNets improves accuracy about $10\%-14\%$ than on GRU\cite{gru} and LSTM\cite{lstm} on average. All these results show the effectiveness of UMSNet.

\subsection{MHEALTH}


MHEALTH dataset has four kinds of sensors, i.e., acceleration sensor, angular velocity sensor, magnetometer, and electrocardiograph sensor. Acceleration sensors, angular velocity sensors, and magnetometers have x, y, and z axes, which we use as 3 channels. The electrocardiograph sensor has 2 leads, which we use as 2 channels. It is worth noting that even if the format of Sensor data is different, we can also customize the UMSNet by using the proposed sensor LRS module. The evaluation methods for MHEALTH dataset are similar to that of the HHAR dataset. Experimental results are presented in Table 4, Figure 5, and Table 5.

\begin{table*}[!htb]
    \begin{center}
        
    \begin{tabular}{|c|c|c|c|c|c|c|}
    \hline
      \textbf{ Method} &\multicolumn{2}{|c|}{$K=6$} & \multicolumn{2}{|c|}{$K=12$}  &  \multicolumn{2}{|c|}{$K=24$}    \\ \cline{2-7}
        \textbf{comparison} & \textbf{Accuracy} & \textbf{Macro-F1} & \textbf{Accuracy} & \textbf{Macro-F1} & \textbf{Accuracy} & \textbf{Macro-F1}  \\ \hline\hline
        GRU\cite{gru} & 0.878 & 0.876 & 0.9337 & 0.933 & 0.9187 & 0.9203  \\ \hline
        LSTM\cite{lstm} & 0.8455 & 0.8417 & 0.9145 & 0.9142 & 0.9346 & 0.9341  \\ \hline
        ResNet\cite{regnet} & 0.9707 & 0.9706 & 0.9701 & 0.9701 & 0.9714 & 0.9713  \\ \hline
        Efficientnet\cite{efficientnet} & 0.9783 & 0.9782 & 0.9714 & 0.9713 & 0.9736 & 0.9735  \\ \hline
        Regnet\cite{regnet} & 0.9785 & 0.9785 & 0.9813 & 0.9814 & 0.9836 & 0.9836  \\ \hline
        UMSNet-A(Ours) & 0.9797 & 0.9798 & 0.983 & 0.9829 & 0.9852 & 0.9851  \\ \hline
        UMSNet-B(Ours) & 0.9802 & 0.9802 & 0.9842 & 0.9842 & 0.9857 & 0.9856  \\ \hline
        UMSNet-C(Ours) & \textbf{0.9804} & \textbf{0.9804} & \textbf{0.9844} & \textbf{0.9844} & \textbf{0.9872} & \textbf{0.9872}  \\ \hline
    \end{tabular}
    \end{center}
    \caption{Performance of different methods on MHEALTH dataset. $K$ is the slice number of a time series data.}
\end{table*}

\begin{figure}[!htb]
\begin{center}

\includegraphics[height=8cm]{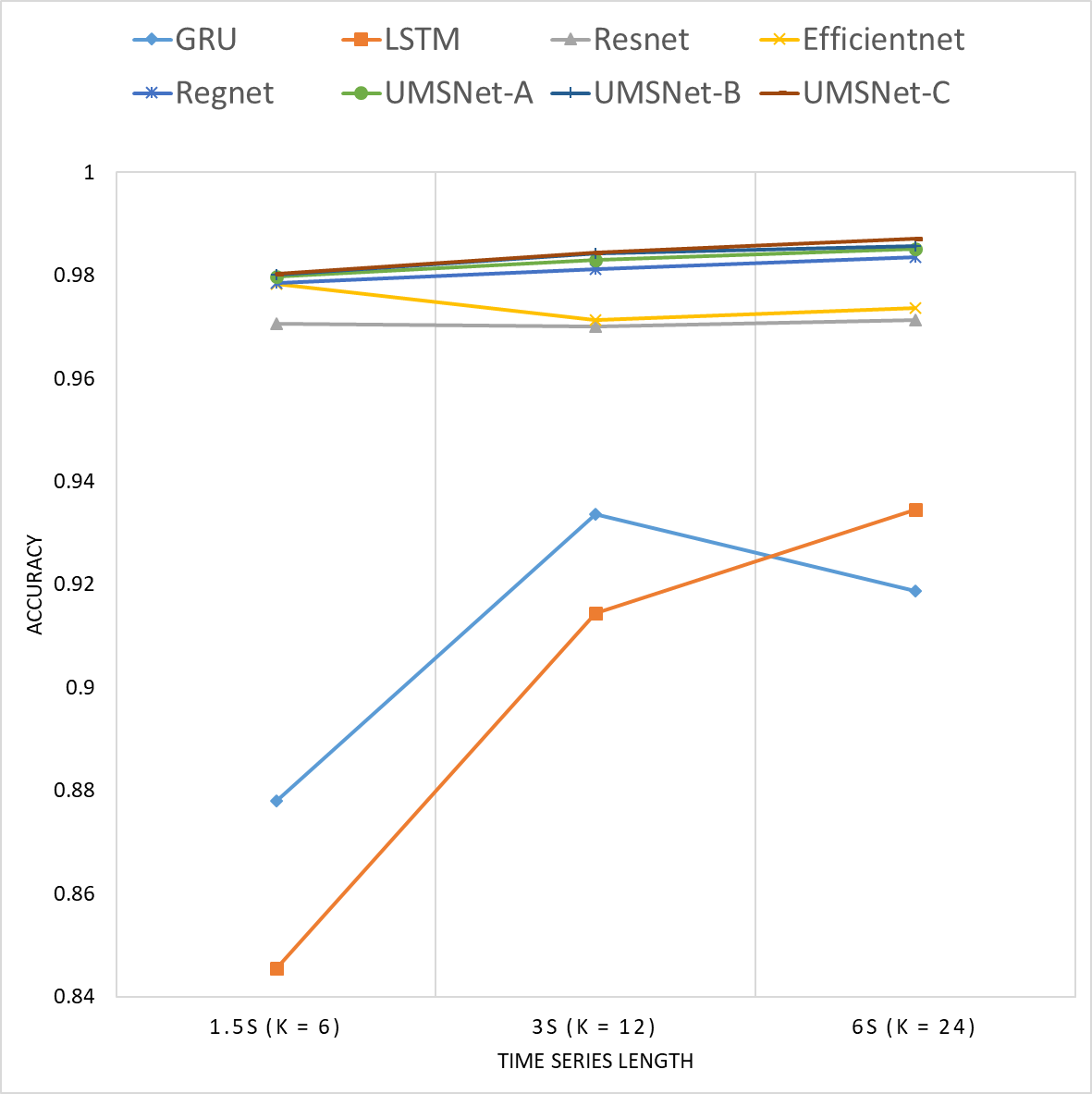}

\end{center}
    \caption{Accuracy of different methods of three time series length setting on MHEALTH dataset.}
\label{fig:short}
\end{figure}

\begin{table}[!htb]
    \begin{center}
        
    \begin{tabular}{|c|c|c|c|}
    \hline
         \textbf{Method} &  \textbf{Params} &  \textbf{Multi-Adds} &  \textbf{Time(ms)} 
         \\ \hline \hline
        GRU\cite{gru} & 28704 & 28560 & 2.958
 \\ \hline
        LSTM\cite{lstm} & 38272 & 38080 & 4.082
\\ \hline
        ResNet\cite{resnet} & 25.557M & 4.089G & 90.996
 \\ \hline
        Efficientnet\cite{efficientnet} & 7.794M & 569.682M & 75.981
 \\ \hline
        Regnet\cite{regnet} & 54.278M & 15.940G  & 168.266
\\ \hline
        UMSNet-A & 5.555M & 36.056M  & 49.565
\\ \hline
        UMSNet-B & 8.340M & 39.550M & 68.284
 \\ \hline
        UMSNet-C & 11.955M & 45.307M & 119.418
 \\ \hline
    \end{tabular}
    \end{center}
    \caption{Comparison of network parameter, computational cost, and the speed of processing a time series of 6 seconds on MHEALTH dataset.}
\end{table}

It can be observed from results that UMSNet-C achieves excellent performance in the MHEALTH dataset. Because MHEALTH dataset has more kinds of sensors than HHAR dataset, all methods improve the accuracy of recognition to a certain extent. However, it can be seen from Figure 5 that our methods still maintains robustness. Considering the network performance and recognition rate comprehensively in Table 4 and Table 5, our overall performance is better than other methods.

Experimental results on these two datasets show that the proposed UMSNet has good generalization and performance for multiple sensor time series.  

\section{Conclusion}

In this paper, we propose the UMSNet framework for human activity recognition.  The UMSNet integrates the fusion residual network and Transformer structure to automatically extract local, global, and time series relationships from multimodal sensors to effectively perform classification tasks. We evaluated the UMSNet using two representative human activity datasets (i.e. HHAR, MHEALTH), in which the UMSNet outperforms other state-of-the-art networks. UMSNet is smaller in network complexity than other networks, requiring very few computing resources to train the network. We have also designed three UMSNet architectures to provide experience for broad application and further adaptation and customization of the framework. In the future, we intend to extend this work to other multi-modal tasks, as well as to use other (larger) datasets for evaluation.

{\small
\bibliographystyle{ieee}
\bibliography{egpaper_for_review}
}

\end{document}